\documentclass[letterpaper]{article} 
\usepackage{aaai2026}  
\usepackage{times}  
\usepackage{helvet}  
\usepackage{courier}  
\usepackage[hyphens]{url}  
\usepackage{graphicx} 
\usepackage{amsmath}
\usepackage{amssymb}
\usepackage{natbib}  
\usepackage{caption} 
\usepackage{algorithm}
\usepackage{algorithmic}

\usepackage{newfloat}
\usepackage{listings}
\DeclareCaptionStyle{ruled}{labelfont=normalfont,labelsep=colon,strut=off} 
\floatstyle{ruled}
\newfloat{listing}{tb}{lst}{}
\floatname{listing}{Listing}
\title{GES-UniGrasp: A Two-Stage Dexterous Grasping Strategy \\With Geometry-Based Expert Selection }
\author{
    Fangting Xu\textsuperscript{\rm 1}\thanks{Authors contributed equally: 12325045@zju.edu.cn.} ,
    Jilin Zhu\textsuperscript{\rm 2},
    Xiaoming Gu\textsuperscript{\rm 1},
    Jianzhong Tang\textsuperscript{\rm 3}\thanks{Corresponding author: jztang@zju.edu.cn.} \\
}
\affiliations{

    \textsuperscript{\rm 1}College of Mechanical Engineering, Zhejiang University
    \textsuperscript{\rm 2}Ocean College, Zhejiang University\\
    \textsuperscript{\rm 3}Institute of Advanced Technology, Zhejiang University

}

\usepackage{bibentry}

\begin{document}

\makeatletter
\def\copyright@text{}
\makeatother
\maketitle

\begin{abstract}
Robust and human-like dexterous grasping of general objects is a critical capability for advancing intelligent robotic manipulation in real-world scenarios. However, existing reinforcement learning methods guided by grasp priors often result in unnatural behaviors. In this work, we present \textit{ContactGrasp}, a robotic dexterous pre-grasp and grasp dataset that explicitly accounts for task-relevant wrist orientation and thumb-index pinching coordination. The dataset covers 773 objects in 82 categories, providing a rich foundation for training human-like grasp strategies. Building upon this dataset, we perform geometry-based clustering to group objects by shape, enabling a two-stage Geometry-based Expert Selection (GES) framework that selects among specialized experts for grasping diverse object geometries, thereby enhancing adaptability to diverse shapes and generalization across categories. Our approach demonstrates natural grasp postures and achieves high success rates of 99.4\% and 96.3\% on the train and test sets, respectively, showcasing strong generalization and high-quality grasp execution.
\end{abstract}


\section{Introduction}

Dexterous grasping plays a vital role in enabling fine-grained human-like manipulation for humanoid robots. However, achieving robust and generalizable control remains highly challenging due to the complex kinematics of dexterous hands, variability in object geometry and materials, and diverse task requirements. Current learning-based approaches are typically divided into two paradigms: imitation learning (IL), which leverages human demonstrations but suffers from high data collection costs and limited transferability; and reinforcement learning (RL), which enables autonomous exploration but depends on task-specific reward shaping, hindering generalization. This gap highlights a central research question: How can we effectively combine human priors with RL’s exploratory capacity to develop adaptive reward mechanisms? Addressing this challenge is key to advancing general-purpose dexterous manipulation in real-world scenarios.
\begin{figure}[t]
\centering
\includegraphics[width=0.9\columnwidth]{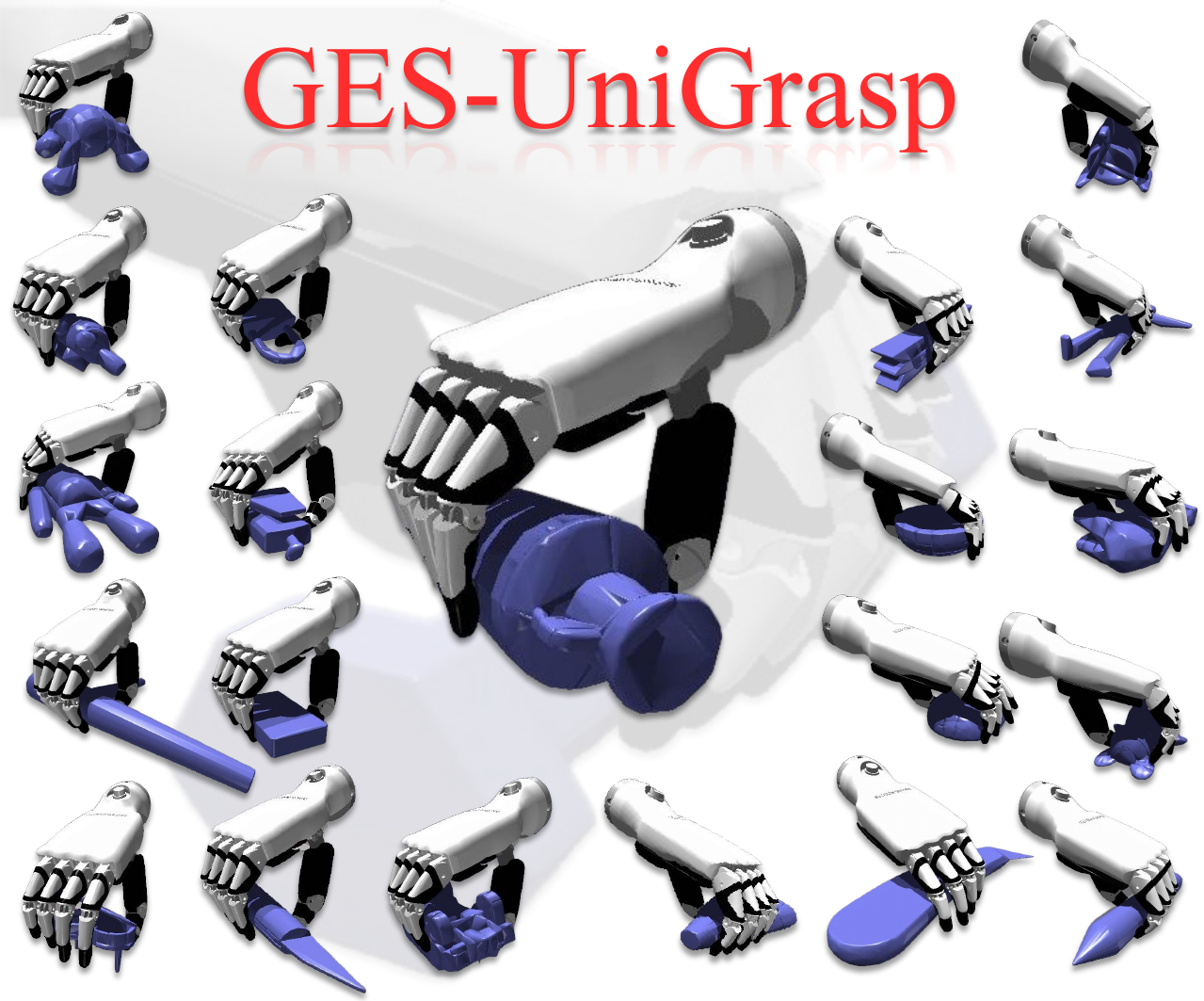} 
\caption{Visualization of our \textit{ContactGrasp} dataset.}
\label{fig1}
\end{figure}

A notable line of work, such as Unidexgrasp\cite{xu2023unidexgrasp}, proposes a goal-conditioned RL framework that uses pre-generated grasp gestures as priors for policy learning. While this improves generalization, it often results in unnatural and constrained grasp poses due to factors such as excessive degrees of freedom—especially in the Shadow Hand causing finger interference—and the diversity or implausibility of target gestures. Moreover, single-policy training typically yields a narrow grasping style lacking behavioral diversity. To address this, Unidexgrasp++\cite{wan2023unidexgrasp++} introduces a generalist-specialist architecture with visual policy distillation, achieving an 11.7\% gain in success rate. However, it still produces non-anthropomorphic poses and suffers from gradient interference in multi-task settings\cite{yu2020gradient}. ResDex\cite{huang2024efficient} adopts a residual Mixture-of-Experts (MOE) model, raising success rates to 88.8\% and improving finger naturalness, but often favors energy-efficient yet suboptimal grasps using only 2–3 fingers. More recently, FunGrasp\cite{huang2025fungrasp} retargets human RGB-D grasp data to robot hands, reaching 75\% real-world success. Yet, its generalization remains limited by the scarcity of demonstrations. Together, these works highlight the need for scalable, biomechanically plausible strategies in dexterous manipulation.

\begin{figure*}[t]
\centering
\includegraphics[width=1.5\columnwidth]{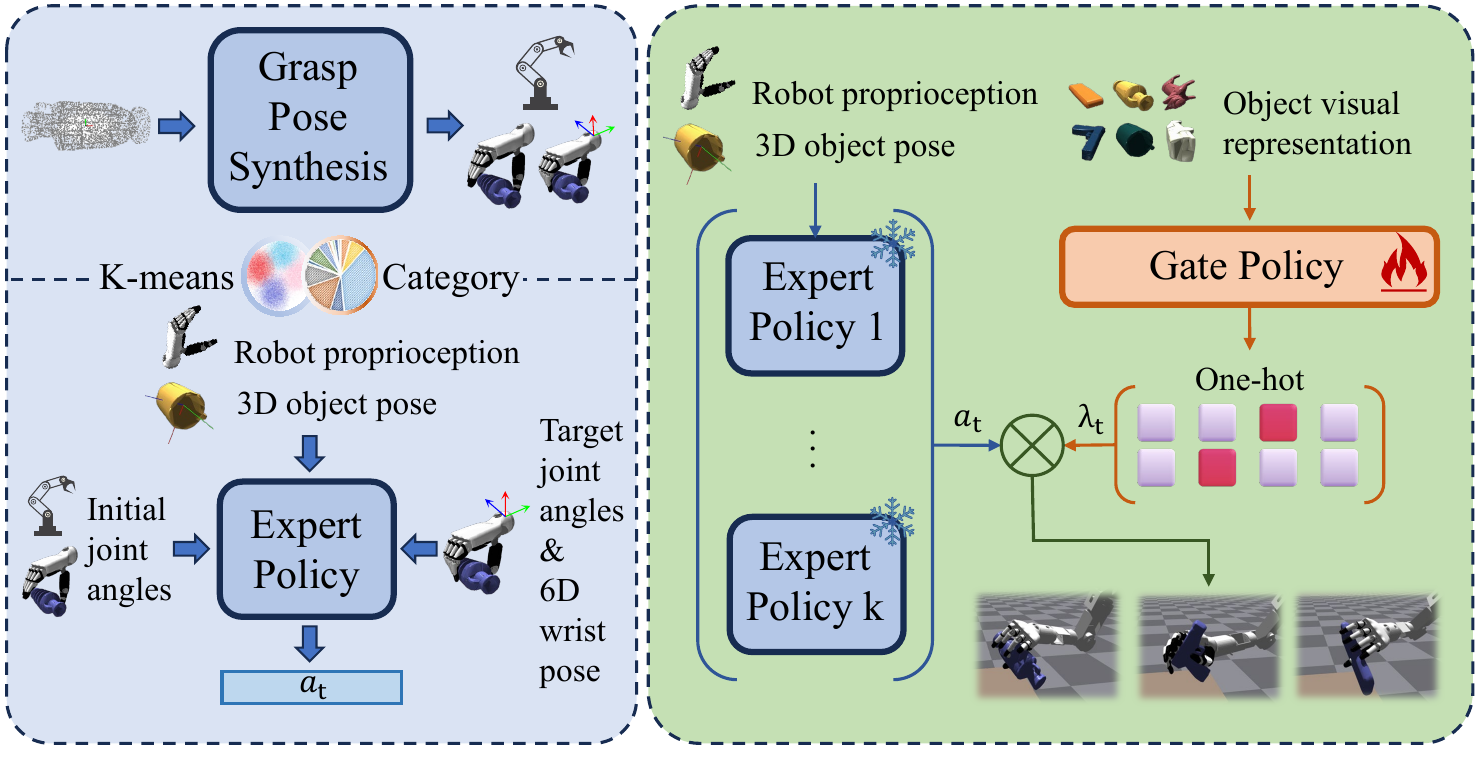}
\caption{Overview of GES-UniGrasp. Our framework consists of a contact-guided grasp synthesis module and a geometry-based multi-expert policy module. It takes object point clouds as input and outputs expert-specific control for grasping.}
\label{fig:overview}
\end{figure*}

To address these challenges, we decompose the dexterous grasping problem into two stages. (1) Contact-guided grasp pose synthesis: This stage automatically generates contact points from object point clouds and synthesizes both pre-grasp and grasp poses that are well-suited for dexterous manipulation tasks. (2) Geometry-based multi-policy grasping: Given a target grasp pose, this stage learns adaptive expert policies and dynamically selects the most appropriate expert based on object geometry to robustly execute the grasp. Both stages present significant challenges, and we make the following key contributions to address them:

To enable efficient grasp synthesis, we propose a contact point generation pipeline and construct a robotic dexterous grasp dataset, \textit{ContactGrasp} (see figure~\ref{fig1}). Our method automatically annotates contact points, significantly improving annotation efficiency. To facilitate unsupervised contact-based grasp synthesis, we introduce a vector retargeting and collision-aware refinement algorithm, which minimizes fingertip-to-contact distances and prevents unrealistic finger–object penetrations. Compared to end-to-end grasp generation methods, contact-based synthesis allows grasp poses to be efficiently retargeted to different dexterous hands by leveraging shared contact point representations, supporting cross-hardware dataset scalability.

To improve learning efficiency and grasp robustness, we adopt a two-stage control strategy: a PD controller guides the hand to a pre-grasp pose, followed by a goal-conditioned RL policy that executes the grasp. Furthermore, to enhance generalization and adaptability, we propose a GES framework in which a geometry-based gating network activates appropriate expert policies based on object shape features, supporting   adaptive grasping across diverse object categories. 

Extensive experiments demonstrate that our method achieves strong performance across both stages. In the grasp pose synthesis stage, our approach efficiently generates high-quality, anthropomorphic grasp poses within 5 seconds per object, achieving a success rate of 71\%. In the grasp execution stage, from geometry-driven expert selection to closed-loop control, our framework yields significant improvements in simulation, achieving a 99.4\% grasp success rate on the train set. To promote further research in dexterous manipulation, we will release our code and dataset.

\section{Related Work}

\subsection{Dexterous Grasp Datasets}
High-quality dexterous grasp datasets are crucial for data-driven grasp synthesis. Due to the high DoF of robotic hands, manual annotation is impractical, leading to two main alternatives: visual capture and procedural generation. Visual methods (e.g., FreiHAND\cite{zimmermann2019freihand}, HO3D\cite{qian2014realtime}, DexYCB\cite{chao2021dexycb}, ContactPose\cite{brahmbhatt2020contactpose}) provide realistic data via RGB/D and multimodal sensing but are limited by hardware and calibration costs. Procedural approaches (e.g., GraspIt!\cite{miller2004graspit}) estimate grasp stability using force closure\cite{nguyen1988constructing} and grasp wrench space(GWS)\cite{borst2004grasp}, but often lack diversity and realism. DexGraspNet\cite{wang2023dexgraspnet} improves efficiency via differentiable force closure, yet procedural methods remain costly and prone to unnatural grasps. Recent contact-based methods (e.g., contact graphs\cite{brahmbhatt2020contactpose,brahmbhatt2019contactdb} or sparse contact points\cite{shao2020unigrasp,wu2022learning,li2022efficientgrasp}) improve plausibility but are either dataset-constrained or overly simplified. To overcome these issues, we propose a geometry-constrained, optimization-enhanced pipeline that predicts anatomically and physically plausible 3D contact points from object point clouds, significantly improving labeling efficiency.

\subsection{Dexterous Grasp Execution}
High-DoF dexterous grasping is a core challenge in robotic manipulation. Traditional analytic approaches\cite{andrews2013goal,bai2014dexterous,dai2023graspnerf} oversimplify hand-object interactions and fail in complex settings. Data-driven methods, including imitation learning with generative models (e.g., CVAE\cite{ye2023learning}, diffusion\cite{pan2024vision}, visual-language-action (VLA)\cite{gbagbe2024bi,fu2024mobile}) and RL, offer improved adaptability but face trade-offs: imitation learning requires large, labeled datasets and object priors, while RL struggles with sparse rewards and poor generalization. Hybrid methods (e.g., inverse RL\cite{christen2022d,mandikal2021learning,rajeswaran2017learning,she2022learning,wu2023learning,nagabandi2020deep,agarwal2023dexterous}) aim to balance both. UniDexGrasp++ improves grasp success via generalist-specialist training, but incurs high training cost and overfitting risk. The MoE framework\cite{jacobs1991adaptive} recently successful in multi-task RL\cite{peng2019mcp}, offers a promising alternative. ResGrasp integrates MoE with residual learning, preserving the strengths of each base policy to train more adaptable grasping strategies, significantly improving efficiency. Building on this, we propose a GES framework that trains expert policies with diverse geometric preferences and dynamically selects the most suitable expert based on the object's shape. This approach maintains stylistic diversity across experts while reducing the computational overhead of multi-policy fusion, achieving both high success rates and strong generalization across object categories.
\begin{figure}[t]
\centering
\includegraphics[width=0.9\columnwidth]{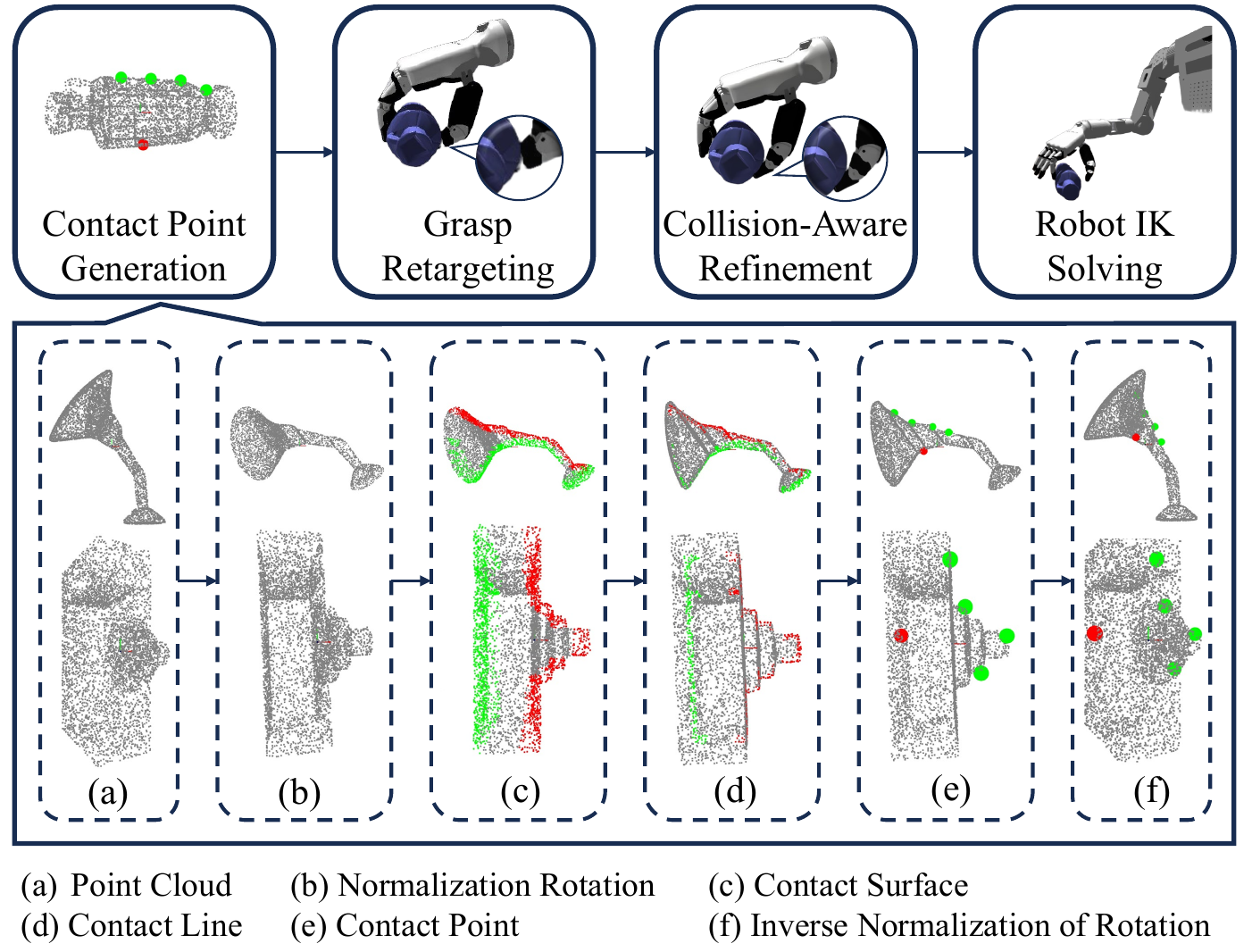}
\caption{Workflow of Contact-Guided Grasp Pose Synthesis.}
\label{fig:workflow and visualize}
\end{figure}

\section{Method}
We propose \textbf{GES-UniGrasp}, a two-stage dexterous grasping framework designed for generalizable object manipulation. The framework comprises: (1) contact-guided grasp pose synthesis, and (2) geometry-based multi-policy grasp execution. An overview is shown in figure~\ref{fig:overview}.

\subsection{Problem Settings and Overview}

The first stage, \textit{contact-guided grasp pose synthesis}, takes the object point cloud $\mathcal{P}_o \in \mathbb{R}^{N \times 3}$ and the table point cloud $\mathcal{P}_t$ as input, generates a set of five-finger contact points $C = \{c_i\}_{i=1}^5$ and a grasp pose $g = (R, t, q)$, where $R \in \mathrm{SO}(3)$, $t \in \mathbb{R}^3$, and $q \in \mathbb{R}^K$ denote the wrist orientation, translation, and joint angles of the dexterous hand. A pre-grasp pose $g_{\text{init}}$ is further derived from $g$ via a consistency transformation.

In the second stage, \textit{geometry-based multi-policy execution}, the system first drives the hand to $g_{\text{init}}$ via trajectory planning. Then, a geometry-based gating network $\pi_\psi^G(\mathcal{P}_o): \mathbb{R}^{N \times 3} \rightarrow \mathbb{R}^k$ selects among $k$ expert grasping policies $\{\pi_{\psi_i}^E\}_{i=1}^{k}$ trained on object clusters formed via K-means and category-based clustering. Experts are trained via progressive curriculum learning, and grasps are considered successful if the object reaches its target position within a threshold, i.e., $\|p_{\text{obj}} - p_{\text{target}}\|_2 < \delta$.

\begin{figure*}[t]
\centering
\includegraphics[width=1.9\columnwidth]{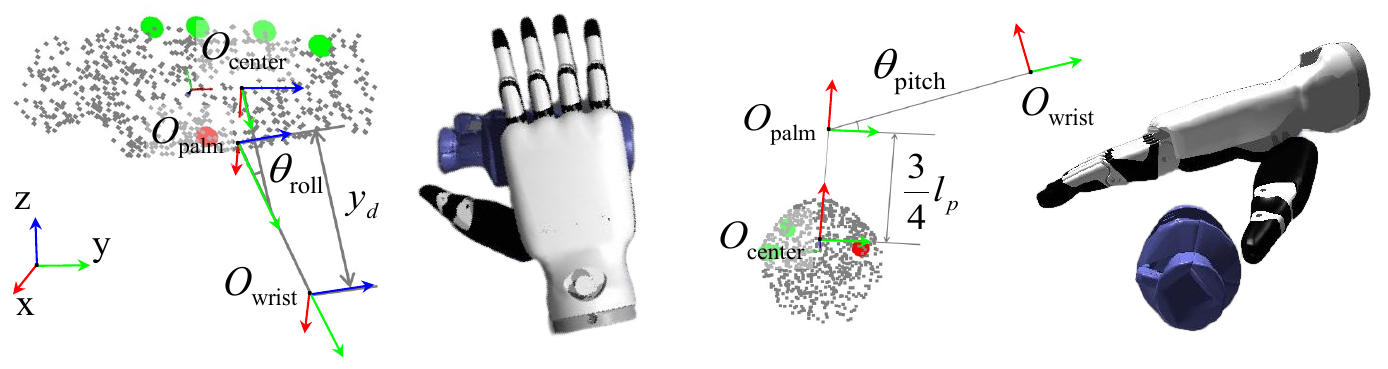}
\caption{Wrist frame transformation based on contact points for Grasp Retargeting.}
\label{fig:wrist transformation}
\end{figure*}

\subsection{Contact-Guided Grasp Pose Synthesis}
Inspired by human biomechanics, we design a four-stage pipeline: contact point generation, grasp retargeting, collision-aware refinement, and robot inverse kinematics (IK) solving (see figure~\ref{fig:workflow and visualize}).

\noindent\textbf{Contact Point Generation} A rule-based pipeline is developed to determine grasp directions and generate five-finger contact configurations. (1) Strategy Selection: Principal Component Analysis (PCA) is appied to the tabletop point cloud. If the principal axis aligns with the z-axis, a horizontal grasp is selected; otherwise, a vertical grasp is used. (2) Normalization Rotation: The object is aligned to a canonical orientation. Full PCA alignment is adopted for non-cylindrical objects, whereas for cylindrical objects, rotation is restricted to the z-axis to preserve their inherent position along the axis of symmetry. (3) Contact Computation: Based on the grasp direction, horizontal slicing and extremum analysis identify potential contact surfaces. Final contact points are refined via local optimization to ensure force closure and five-finger coordination.

\noindent\textbf{Grasp Retargeting}  
We extract four grasp topology descriptors $G = (\mathcal{F}, \mathbf{n}_{\text{palm}}, \mathbf{d}_{\text{palm}}, \mathbf{u}_{\text{palm}})$: (i) contact point set $\mathcal{F}$; (ii) palm normal $\mathbf{n}_{\text{palm}}$ via singular value decomposition(SVD) computed on $\mathcal{F}$; (iii) grasp axis $\mathbf{d}_{\text{palm}}$ (wrist-to-middle); and (iv) lateral axis $\mathbf{u}_{\text{palm}}$ (middle-to-little). A consistent pronation–pitch coupling is modeled with $\theta_{\text{roll}}=10^\circ$, $\theta_{\text{pitch}}=20^\circ$, aligning grasp orientation with the object axis.

The wrist origin is computed as:
\begin{align}
O_{\text{palm}} &= O_{\text{center}} + [0, 0, 3l_p/4]^\top \\
O_{\text{wrist}} &= O_{\text{palm}} + y_d \cdot R_{\text{wrist}} \cdot \mathbf{y}_{\text{wrist}}
\end{align}

Where $O_{\text{center}}$, $O_{\text{palm}}$ is the contact-derived geometric center and the palm center, $l_p$ is the anatomical length of the middle finger. The wrist offset is defined along the wrist frame’s $y$-axis by a scalar $y_d$ and its unit vector $\mathbf{y}_{\text{wrist}}$, and $R_{\text{wrist}}$ is the wrist-to-world rotation matrix. To enhance the clarity of visualization, the wrist frame $ \sum _ \text{wrist} $ is illustrated in figure~\ref{fig:wrist transformation} from two different perspectives.


All contact points are transformed to the wrist frame. Retargeting is then formulated as:
\begin{equation}
\begin{aligned}
\min_{q_t} \quad & \sum_{i=0}^{N} \left\| \alpha v_t^i - f_i(q_t) \right\|^2 + \beta \left\| q_t - q_{t-1} \right\|^2 \\
\text{s.t.} \quad & q_l \leq q_t \leq q_u
\end{aligned}
\end{equation}

Here $q_t$ is the robot hand joint angles at timestep $t$, $v_t^i$ the $i$-th keypoint of the human hand, and $f_i(\cdot)$ the robot’s forward kinematics. $\alpha$ scales between hand sizes, and $\beta$ ensures temporal smoothness.

\noindent\textbf{Collision-Aware Refinement}  
To eliminate interpenetration, we perform physics-based refinement using Isaac Gym\cite{makoviychuk2021isaac}. Given desired joint positions $q_d^{(i)}$ from retargeting, the PD controller applies:
\begin{equation}
\tau^{(i)} = K_p(q_d^{(i)} - q^{(i)}) + K_d(\dot{q}_d^{(i)} - \dot{q}^{(i)})
\end{equation}

When fingertip contact exceeds a threshold $\|f_i\|_2 > \gamma$, we freeze the joint target: $q_d^{(i)}(t^+) = q^{(i)}(t^-)$. This closed-loop feedback prevents penetration and stabilizes the final pose $g$.

A final consistency transformation shifts $g$ 0.02m along the approach direction and resets intermediate joints to produce the pre-grasp pose $g_{\text{init}}$.

\subsection{Geometry-Based Multi-Policy Grasping}

\noindent\textbf{Two-Stage Control Framework}  
We adopt a two-stage grasping framework that separates manipulation into:  
(1) \textit{Pre-grasp Reaching}: Given synthesized grasp and pre-grasp poses, the robot arm tracks these via inverse kinematics and trajectory planning for efficient positioning.  
(2) \textit{Grasp Execution}: A goal-conditioned RL policy is trained to perform robust, closed-loop grasping without visual feedback, starting from the pre-grasp pose. This staged design confines RL exploration to a well-initialized state space, enhancing sample efficiency and robustness. Upon failure, the system can replan and retry, forming a closed-loop grasp pipeline.

\noindent\textbf{Expert Grasping Policy}  
Expert policies are trained using Proximal Policy Optimization (PPO) on a humanoid robot, with only the right arm and dexterous hand joints active. The observation includes:
\begin{align}
s &= (J,\ a_{t-1},\ p_o,\ G) \\
J &= (q,\ \dot{q},\ c_b,\ f,\ \tilde{p}_f,\ T_{\text{wrist}},\ \dot{T}_{\text{wrist}})
\end{align}
where $q$, $\dot{q}$ are joint positions, velocities, $c_b$, $f$ are contact states and fingertip forces, $a_{t-1}$ is the previous action, $\tilde{p}_f$ is fingertip position in wrist frame, $T_{\text{wrist}}$ and $\dot{T}_{\text{wrist}}$ are the wrist’s 6D pose and velocity. $p_o$ is 3D object position, and $G = (\widehat{g}_p,\ \widehat{g}_r,\ g_q)$ is the pose difference between current and target, including wrist position, orientation, and joint angles. The policy outputs desired joint angles $a_t$ for both hand and arm.

Inspired by UniDexGrasp, we design a refined reward function with the following components:
$ r = \omega_{g} r_{g} + \omega_{r} r_{r} + \omega_{l} r_{l} + \omega_{m} r_{m} + \omega_{s} r_{s} + \omega_{c} r_{c} $. It comprises goal reward $r_{g}$, reach reward $r_{r}$, lift reward $r_{l}$, move reward $r_{m}$ , smooth reward $r_{s}$, and consistency reward $r_{c}$.  

First, we reformulate the two proximity-related terms—the distance between the hand and the object $r_{r}$ and the distance between the object and the target position $r_{l}$—as exponentially decaying functions: $\exp(-\lambda r)$ where \(r\) denotes the original reward, to more effectively guide the system to grasp successfully.
Second, to prevent unnatural finger extensions caused by redundant degrees of freedom, we add per-finger joint angle constraints in the \(r_{g}\) term: $r_{pf} = \max_f \left\| \Delta q_f \right\|_1$,
where \(\Delta q_f\) is the joint angle difference for finger \(f\).
Third, to reduce high-frequency oscillations in control signals, we introduce a joint velocity penalty: $r_{s} = \sum_{i \in \text{joints}} v_i^2$.
Finally, to encourage coordinated opposition and improve grasp stability through learned synergies, we add a directional consistency constraint: $r_{c} = \sum_{j} \operatorname{Var}(\operatorname{sign}(v_j))$, where \(j\) indexes the same-position joints across the four fingers.

\noindent\textbf{Geometry-Based Expert Selection Architecture} 
At the architecture level, we first cluster objects into geometry-similar groups using K-means and category-based clustering. Based on this, we adopt a three-stage curriculum training strategy:

(1) \emph{Intra-cluster central training}: each expert is trained on objects near the geometric center of a cluster;  
(2) \emph{Intra-cluster expansion}: training gradually expands to all objects within the cluster;  
(3) \emph{Inter-cluster generalization}: the expert is fine-tuned on the full object set.

This process yields expert policies tailored to specific geometric distributions while preserving generalization. For objects, we identify the top two experts by performance and collect failure cases where both perform poorly. These hard cases often share common geometric challenges (e.g., lacking pinch surfaces or large centroid shifts). We then train a third expert on these failure cases using only the first two curriculum stages, improving robustness.

At the policy fusion level, we design a Geometry-based gating network that takes object point cloud features as input and predicts a softmax distribution over expert success rates. The expert with the highest predicted score is selected for execution.

Given the similarity to point cloud classification, we use CurveNet~\cite{xiang2021walk} as the backbone and optimize the gating network with the Kullback–Leibler divergence:
\begin{equation}
\mathcal{L}_{\mathrm{KL}} = \frac{1}{N} \sum_{i=1}^{N} D_{\mathrm{KL}}(P_i \,\|\, \hat{P}_i),
\end{equation}
where \(P_i\) is the target distribution (normalized success rates) and \(\hat{P}_i\) is the predicted distribution. This loss encourages the gating network to model expert performance in a geometry-sensitive way.

\begin{table}[t]
\centering
\begin{tabular}{c|c|c|c}
\hline
 & \textbf{Contact Point} & \textbf{Grasp} & \textbf{Collision} \\
 & \textbf{Generation} & \textbf{Retargeting} & \textbf{Refinement} \\
 \hline
SR(\%)& 87.7 & 88.9 & 91.5 \\
T/Obj.(s)& 0.15±0.01 & 0.03±0.01 & 5.15±0.81 \\
\hline
\end{tabular}
\caption{Success rate (SR) and average generation time per object (T/obj.) for the grasp pose synthesis module.}
\label{table1}
\end{table}

\begin{figure}[t]
\centering
\includegraphics[width=0.9\columnwidth]{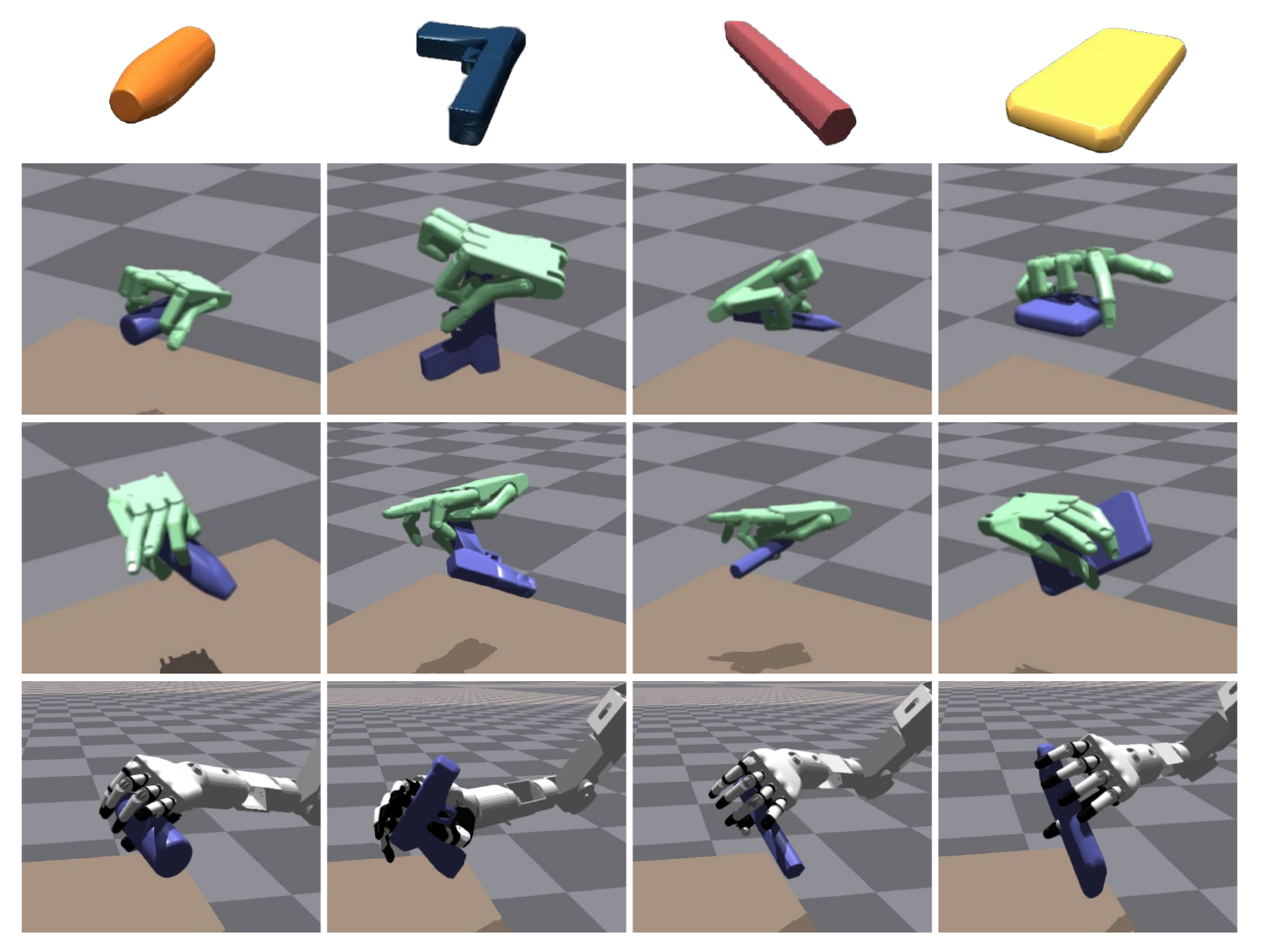}
\caption{Visualization of the three grasping strategies applied to four different objects. From top to bottom, the rows correspond to UniDexGrasp, ResDex, and GES-UniGrasp, respectively.
}
\label{fig:grasp gesture compare}
\end{figure}

\begin{table}[t]
\centering
\begin{tabular}{c|c|cc}
\hline
 & & \multicolumn{2}{c}{\textbf{Test(\%)}} \\
\textbf{Methods} & \textbf{Train(\%)} & \textbf{Uns.Obj} & \textbf{Uns.Obj} \\
  & & \textbf{Seen Cat.} & \textbf{Uns.Cat.} \\
\hline
Unidexgrasp & 79.4 & 74.3 & 70.8 \\
Unidexgrasp++ & 87.9 & 84.3 & 83.1 \\
ResDex & 94.6 & 94.4 & 95.4 \\
\hline
Expert1 & 93.9 & 93.9 & 89.1 \\
Expert2 & 95.4 & 93.0 & 93.4 \\ 
Expert3 & 88.8 & 92.0 & 83.8 \\ 
\hline
GES-UniGrasp & \textbf{99.4} & \textbf{96.3} & \textbf{96.3} \\ 
\hline
\end{tabular}
\caption{Success rates of state-based policies.}
\label{table2}
\end{table}

\begin{figure}[t]
\centering
\includegraphics[width=0.7\columnwidth]{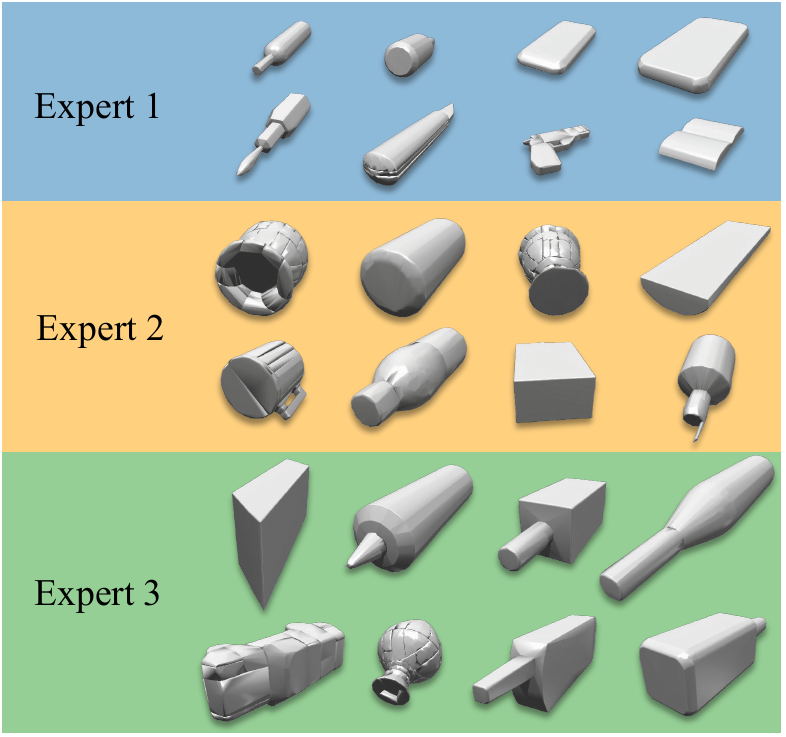}
\caption{Geometric preferences of different expert grasping strategies.}
\label{fig:Geometric preferences}
\end{figure}

\section{Experiment}
\subsection{Environment Setup and Data}
To evaluate the effectiveness of the \textit{ContactGrasp} dataset and the proposed GES-UniGrasp framework, we construct a parallel simulation environment in Isaac Gym, based on a humanoid robotic platform. Each trial is initialized from a pre-grasp pose, and is considered successful if the robot can lift the object to a height of 0.2 meters above the table and maintain that position up to 200 simulation timesteps. The dataset used to train the gating network is generated from the expert policies, with the training and testing splits aligned with those used in expert policy training. From a computational standpoint, contact point generation and retargeting are executed on a CPU, while all other components are deployed across two NVIDIA A100 GPUs.To benchmark performance, we compare GES-UniGrasp with several state-of-the-art methods, including UniDexGrasp, UniDexGrasp++, and ResDex.

\noindent\textbf{Statistics} The \textit{ContactGrasp} dataset consists of high-quality grasp annotations for 773 object instances across 82 categories. These instances are split into three subsets: 563 for training, 132 from seen categories but with unseen instances, and 78 from entirely unseen categories.

\subsection{Grasp Pose Synthesis}
DexGraspNet\cite{wang2023dexgraspnet} focuses on generating a diverse range of grasp poses but fails to explicitly address collisions with supporting surfaces and frequently yields unnatural  configurations as a result of its programmatic generation approach. In comparison, the \textit{ContactGrasp} dataset is designed to efficiently generate consistent and human-like grasps, with a particular focus on natural thumb-to-finger pinch patterns that reflect real-world grasping behaviors.

\begin{figure}[t]
\centering
\includegraphics[width=0.9\columnwidth]{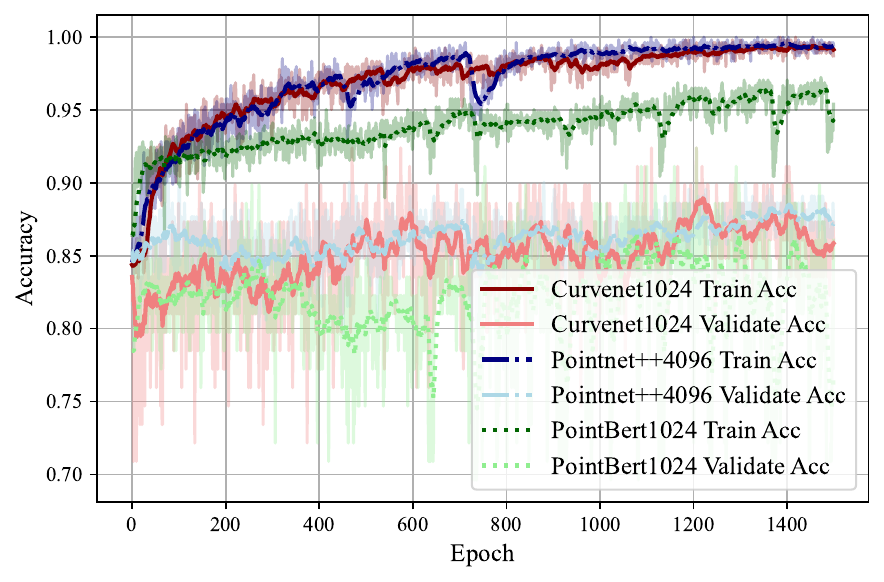}
\caption{Gate network training and validation accuracy curves across different backbone architectures.}
\label{fig:Gate network training}
\end{figure}

\begin{figure}[t]
\centering
\includegraphics[width=0.9\columnwidth]{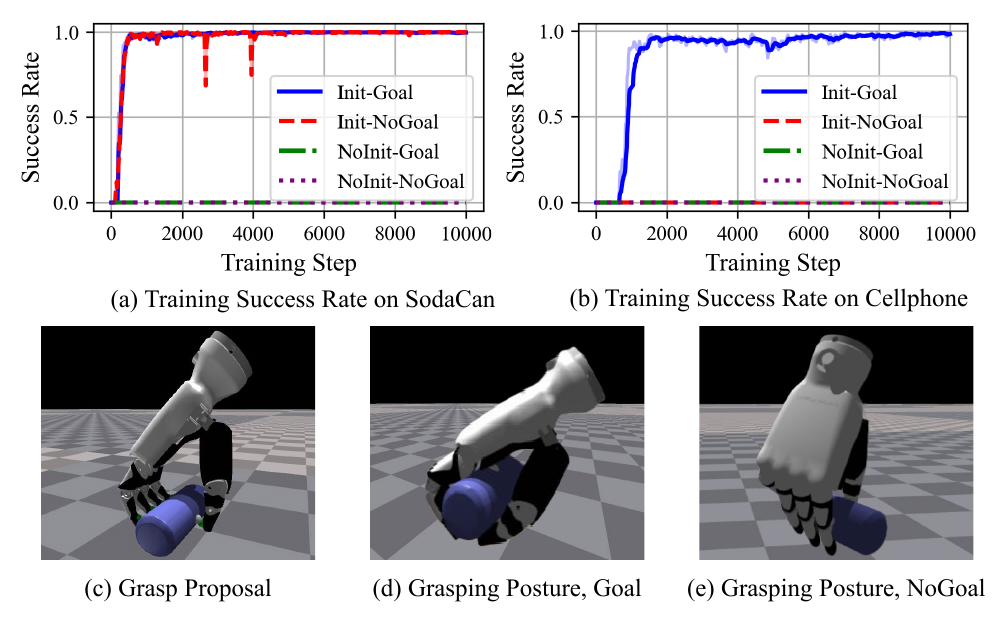}
\caption{Ablation study on the effects of pre-grasp and grasp pose priors.}
\label{fig:ablation}
\end{figure}

While DexGraspNet required approximately 950 hours of NVIDIA A100 GPU time to generate 1.32 million valid grasp poses, our method demonstrates a significantly more resource-efficient alternative. As reported in table~\ref{table1}, our approach generates a grasp pose for each object in just 5 seconds, while maintaining high success rates across the three stages of the grasp synthesis pipeline: 87.7\%, 88.9\%, and 91.5\%, respectively. This efficiency underscores the practicality of our system for scalable data generation.

As illustrated in figure~\ref{fig:grasp gesture compare}, grasp policies trained on DexGraspNet exhibit high diversity in grasp suggestions, but this diversity does not translate into more diverse or effective reinforcement learning policies. In contrast, the more consistent grasp suggestions provided by \textit{ContactGrasp} enable the learned policies to adopt more stable and human-like thumb-index pinch strategies, leading to improved grasp stability and control.

\subsection{Grasp Execution}
\textbf{Main Results} As shown in table~\ref{table2}, our method achieves significant performance gains over baseline approaches. Among state-based policies, the best-performing baseline, ResDex, achieves approximately 95\% success on both the train and test sets. In contrast, our method reaches 99.4\% on the train set and 96.3\% on both test sets. Additionally, the proposed geometric gating network enables the policy to adaptively select the most appropriate expert based on object geometry, further improving performance over single-expert baselines and highlighting the benefits of geometry-based expert selection for universal grasping.

\noindent\textbf{Analysis of the Training Process} To investigate the geometric preferences of individual expert networks, we selected objects for which a single expert policy achieved a grasp success rate exceeding 95\%. As illustrated in figure~\ref{fig:Geometric preferences}, Expert 1 demonstrates superior adaptability to small, flat objects; Expert 2 is more effective with medium-sized geometries and objects exhibiting slight initial rotations; while Expert 3 generalizes better to larger, box-shaped objects, providing stronger pinch forces.

\begin{figure}[t]
\centering
\includegraphics[width=0.9\columnwidth]{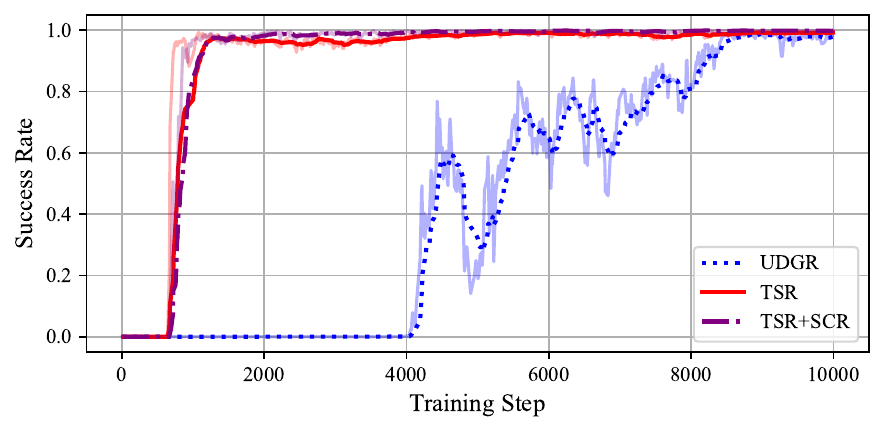}
\caption{Training convergence comparison across three reward formulations.}
\label{fig:Training convergence}
\end{figure}

\begin{table}[t]
\centering
\begin{tabular}{c|c|c|c}
\hline
 & \textbf{Arm Osc.↓} & \textbf{Hand Osc.↓} & \textbf{FDC↑} \\
\hline
UDGR&987&5970&\textbf{-87.8} \\
TSR&3019&92234&-140.37 \\
TSR+SCR&\textbf{347.6}&\textbf{731.9}&-108.5 \\
\hline
\end{tabular}
\caption{Comparison of grasp strategy performance under three different reward formulations, evaluated in terms of arm oscillations(Arm Osc.), hand oscillations(Hand Osc.) and  Finger Directional Consistency(FDC).}
\label{table3}
\end{table}

\begin{figure*}[t]
\centering
\includegraphics[width=1.9\columnwidth]{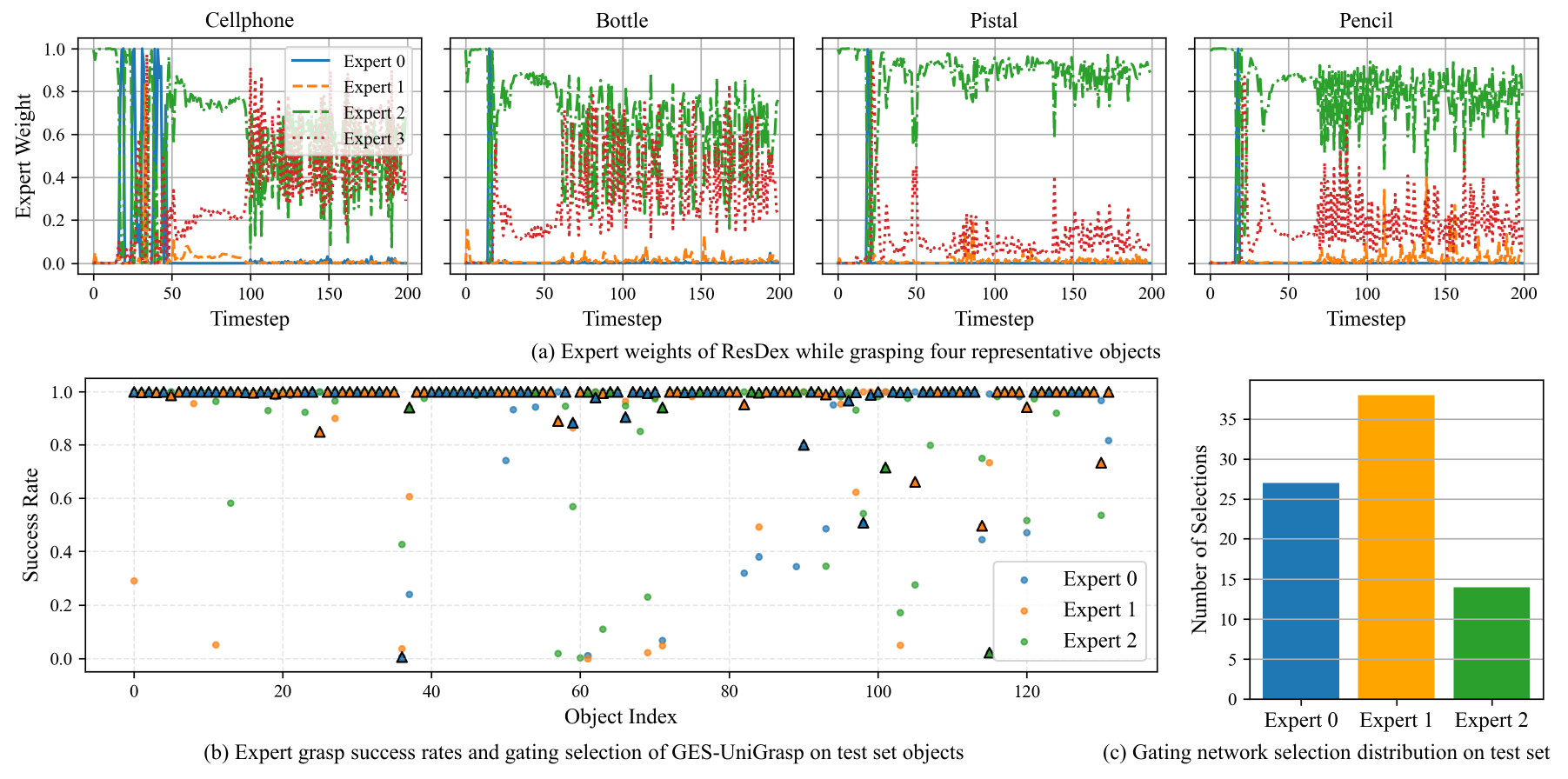}
\caption{Comparison of expert fusion strategies in GES-UniGrasp and ResDex.}
\label{fig:expert fusion}
\end{figure*}

During the training of the gating network, we explored network architectures better suited for geometry-based classification(see figure~\ref{fig:Gate network training}). Point-BERT\cite{yu2022point}, which employs a transformer-based structure, exhibited slower convergence and greater fluctuations in training performance for this task. In contrast, CurveNet\cite{xiang2021walk} and PointNet++\cite{qi2017pointnet++} showed more stable behavior. We also experimented with different point cloud resolutions, using both 4096 and 1024 input points. Results indicate that the model trained with 1024 points achieves better generalization.

Considering both model stability and performance, we ultimately selected the CurveNet-based architecture with 1024-point input as our final gating network. This model achieved grasp success rates of 98.0\% on the train set and 92.4\% on the validation set.

\noindent\textbf{Ablation Experiments} 
We evaluate our method under three configurations to assess the impact of pre-grasp and grasp pose priors:
\begin{itemize}
    \item \textbf{NoInit-NoGoal}: No prior on either the initial hand pose (pre-grasp) or the target grasp pose; the most challenging baseline.
    \item \textbf{Init-NoGoal}: Provides the initial hand pose but not the target grasp pose; evaluates the effect of pre-grasp priors.
    \item \textbf{NoInit-Goal}: Provides the target grasp pose but not the initial hand pose; evaluates the benefit of grasp pose priors.
\end{itemize}

As shown in figure~\ref{fig:ablation}(a-b), the \textbf{Init-Goal} setting (with both pre-grasp and grasp pose priors) consistently achieves fast convergence across objects. In the \textbf{Init-NoGoal} setting, only a subset of objects converge efficiently, indicating that pre-grasp priors alone are sometimes sufficient. In contrast, the \textbf{NoInit} setting struggles with exploration—often displaying ineffective behaviors such as blindly pressing down—and typically fails to converge within 10,000 steps. Visualizations (see in figure~\ref{fig:ablation}(c-e)) further show that the absence of grasp proposals leads to unnatural hand configurations (e.g., fully extended fingers), while their inclusion enables more natural, human-like grasps.

\noindent\textbf{Reward Design Comparison} 
Table~\ref{table3} compares different reward formulations in terms of convergence speed, motion stability, and finger directional consistency. The baseline UniDexGrasp Reward (UDGR) achieves high finger consistency but exhibits slow convergence(see figure~\ref{fig:Training convergence}). Introducing an exponentially decaying Task Success Reward (TSR), which consists of $r_g$, $r_r$, $r_l$, and $r_m$, accelerates learning but leads to noticeable oscillations in both the arm and fingers. By further adding the Smoothness \& Consistency Reward (SCR), composed of $r_s$ and $r_c$, we observe a substantial reduction in motion oscillations and improved directional alignment among fingers.

\noindent\textbf{Expert Fusion Analysis} 
Figure~\ref{fig:expert fusion}(a) shows the expert weight dynamics of ResDex when grasping four representative objects. The results reveal significant expert redundancy, unbalanced usage, and persistent weight oscillations, even after stable grasp configurations are achieved. Despite differences in object geometry, the policy tends to converge to a fixed grasp pattern, indicating limited adaptability and underutilization of expert diversity. In contrast, GES-UniGrasp exhibits better performance and more uniform expert allocation.
Figure~\ref{fig:expert fusion}(b) shows the success rates of each expert on the seen-category test set. Triangular markers indicate the expert selected by the gating network. The results demonstrate that experts exhibit complementary geometric preferences, and the gating network consistently selects the most appropriate expert.
Figure~\ref{fig:expert fusion}(c) presents expert selection distributions on the unseen-category test set, showing that GES-UniGrasp achieves more balanced and effective expert utilization, which contributes to improved grasp performance.

\section{Conclusion}
To enable robust and human-like dexterous grasping of general objects, we present a dexterous grasp dataset \textit{ContactGrasp}, along with a policy-fusion framework GES-UniGrasp. Our key technical contributions include: (1) a novel contact-guided grasp synthesis pipeline, comprising grasp retargeting followed by collision-aware refinement; (2) a two-stage control framework that reduces policy learning complexity and improves convergence stability; (3) a GES framework for adaptive grasp planning across diverse object geometries. We demonstrate that GES-UniGrasp significantly outperforms prior methods on the \textit{ContactGrasp} dataset, particularly in exhibiting anthropomorphic grasp gestures.

Despite these strengths, our approach has several limitations. The grasp synthesis process still relies on handcrafted heuristics. The gating network for expert selection may exhibit limited generalization. Furthermore, the current system has not yet been deployed on physical hardware.

\bibliography{aaai2026}

\begin{thebibliography}{37}
\providecommand{\natexlab}[1]{#1}

\bibitem[{Agarwal et~al.(2023)Agarwal, Uppal, Shaw, and Pathak}]{agarwal2023dexterous}
Agarwal, A.; Uppal, S.; Shaw, K.; and Pathak, D. 2023.
\newblock Dexterous functional grasping.
\newblock \emph{arXiv preprint arXiv:2312.02975}.

\bibitem[{Andrews and Kry(2013)}]{andrews2013goal}
Andrews, S.; and Kry, P.~G. 2013.
\newblock Goal directed multi-finger manipulation: Control policies and analysis.
\newblock \emph{Computers \& Graphics}, 37(7): 830--839.

\bibitem[{Bai and Liu(2014)}]{bai2014dexterous}
Bai, Y.; and Liu, C.~K. 2014.
\newblock Dexterous manipulation using both palm and fingers.
\newblock In \emph{2014 IEEE International Conference on Robotics and Automation (ICRA)}, 1560--1565. IEEE.

\bibitem[{Borst, Fischer, and Hirzinger(2004)}]{borst2004grasp}
Borst, C.; Fischer, M.; and Hirzinger, G. 2004.
\newblock Grasp planning: How to choose a suitable task wrench space.
\newblock In \emph{IEEE International Conference on Robotics and Automation, 2004. Proceedings. ICRA'04. 2004}, volume~1, 319--325. IEEE.

\bibitem[{Brahmbhatt et~al.(2019)Brahmbhatt, Ham, Kemp, and Hays}]{brahmbhatt2019contactdb}
Brahmbhatt, S.; Ham, C.; Kemp, C.~C.; and Hays, J. 2019.
\newblock Contactdb: Analyzing and predicting grasp contact via thermal imaging.
\newblock In \emph{Proceedings of the IEEE/CVF conference on computer vision and pattern recognition}, 8709--8719.

\bibitem[{Brahmbhatt et~al.(2020)Brahmbhatt, Tang, Twigg, Kemp, and Hays}]{brahmbhatt2020contactpose}
Brahmbhatt, S.; Tang, C.; Twigg, C.~D.; Kemp, C.~C.; and Hays, J. 2020.
\newblock ContactPose: A dataset of grasps with object contact and hand pose.
\newblock In \emph{Computer Vision--ECCV 2020: 16th European Conference, Glasgow, UK, August 23--28, 2020, Proceedings, Part XIII 16}, 361--378. Springer.

\bibitem[{Chao et~al.(2021)Chao, Yang, Xiang, Molchanov, Handa, Tremblay, Narang, Van~Wyk, Iqbal, Birchfield et~al.}]{chao2021dexycb}
Chao, Y.-W.; Yang, W.; Xiang, Y.; Molchanov, P.; Handa, A.; Tremblay, J.; Narang, Y.~S.; Van~Wyk, K.; Iqbal, U.; Birchfield, S.; et~al. 2021.
\newblock DexYCB: A benchmark for capturing hand grasping of objects.
\newblock In \emph{Proceedings of the IEEE/CVF conference on computer vision and pattern recognition}, 9044--9053.

\bibitem[{Christen et~al.(2022)Christen, Kocabas, Aksan, Hwangbo, Song, and Hilliges}]{christen2022d}
Christen, S.; Kocabas, M.; Aksan, E.; Hwangbo, J.; Song, J.; and Hilliges, O. 2022.
\newblock D-grasp: Physically plausible dynamic grasp synthesis for hand-object interactions.
\newblock In \emph{Proceedings of the IEEE/CVF Conference on Computer Vision and Pattern Recognition}, 20577--20586.

\bibitem[{Dai et~al.(2023)Dai, Zhu, Geng, Ruan, Zhang, and Wang}]{dai2023graspnerf}
Dai, Q.; Zhu, Y.; Geng, Y.; Ruan, C.; Zhang, J.; and Wang, H. 2023.
\newblock Graspnerf: Multiview-based 6-dof grasp detection for transparent and specular objects using generalizable nerf.
\newblock In \emph{2023 IEEE International Conference on Robotics and Automation (ICRA)}, 1757--1763. IEEE.

\bibitem[{Fu, Zhao, and Finn(2024)}]{fu2024mobile}
Fu, Z.; Zhao, T.~Z.; and Finn, C. 2024.
\newblock Mobile aloha: Learning bimanual mobile manipulation with low-cost whole-body teleoperation.
\newblock \emph{arXiv preprint arXiv:2401.02117}.

\bibitem[{Gbagbe et~al.(2024)Gbagbe, Cabrera, Alabbas, Alyunes, Lykov, and Tsetserukou}]{gbagbe2024bi}
Gbagbe, K.~F.; Cabrera, M.~A.; Alabbas, A.; Alyunes, O.; Lykov, A.; and Tsetserukou, D. 2024.
\newblock Bi-vla: Vision-language-action model-based system for bimanual robotic dexterous manipulations.
\newblock In \emph{2024 IEEE International Conference on Systems, Man, and Cybernetics (SMC)}, 2864--2869. IEEE.

\bibitem[{Huang et~al.(2025)Huang, Zhang, Wu, Christen, and Song}]{huang2025fungrasp}
Huang, L.; Zhang, H.; Wu, Z.; Christen, S.; and Song, J. 2025.
\newblock Fungrasp: Functional grasping for diverse dexterous hands.
\newblock \emph{IEEE Robotics and Automation Letters}.

\bibitem[{Huang et~al.(2024)Huang, Yuan, Fu, and Lu}]{huang2024efficient}
Huang, Z.; Yuan, H.; Fu, Y.; and Lu, Z. 2024.
\newblock Efficient Residual Learning with Mixture-of-Experts for Universal Dexterous Grasping.
\newblock \emph{arXiv preprint arXiv:2410.02475}.

\bibitem[{Jacobs et~al.(1991)Jacobs, Jordan, Nowlan, and Hinton}]{jacobs1991adaptive}
Jacobs, R.~A.; Jordan, M.~I.; Nowlan, S.~J.; and Hinton, G.~E. 1991.
\newblock Adaptive mixtures of local experts.
\newblock \emph{Neural computation}, 3(1): 79--87.

\bibitem[{Li et~al.(2022)Li, Baron, Zhang, and Rojas}]{li2022efficientgrasp}
Li, K.; Baron, N.; Zhang, X.; and Rojas, N. 2022.
\newblock Efficientgrasp: A unified data-efficient learning to grasp method for multi-fingered robot hands.
\newblock \emph{IEEE Robotics and Automation Letters}, 7(4): 8619--8626.

\bibitem[{Makoviychuk et~al.(2021)Makoviychuk, Wawrzyniak, Guo, Lu, Storey, Macklin, Hoeller, Rudin, Allshire, Handa et~al.}]{makoviychuk2021isaac}
Makoviychuk, V.; Wawrzyniak, L.; Guo, Y.; Lu, M.; Storey, K.; Macklin, M.; Hoeller, D.; Rudin, N.; Allshire, A.; Handa, A.; et~al. 2021.
\newblock Isaac gym: High performance gpu-based physics simulation for robot learning.
\newblock \emph{arXiv preprint arXiv:2108.10470}.

\bibitem[{Mandikal and Grauman(2021)}]{mandikal2021learning}
Mandikal, P.; and Grauman, K. 2021.
\newblock Learning dexterous grasping with object-centric visual affordances.
\newblock In \emph{2021 IEEE international conference on robotics and automation (ICRA)}, 6169--6176. IEEE.

\bibitem[{Miller and Allen(2004)}]{miller2004graspit}
Miller, A.~T.; and Allen, P.~K. 2004.
\newblock Graspit! a versatile simulator for robotic grasping.
\newblock \emph{IEEE Robotics \& Automation Magazine}, 11(4): 110--122.

\bibitem[{Nagabandi et~al.(2020)Nagabandi, Konolige, Levine, and Kumar}]{nagabandi2020deep}
Nagabandi, A.; Konolige, K.; Levine, S.; and Kumar, V. 2020.
\newblock Deep dynamics models for learning dexterous manipulation.
\newblock In \emph{Conference on robot learning}, 1101--1112. PMLR.

\bibitem[{Nguyen(1988)}]{nguyen1988constructing}
Nguyen, V.-D. 1988.
\newblock Constructing force-closure grasps.
\newblock \emph{The International Journal of Robotics Research}, 7(3): 3--16.

\bibitem[{Pan, Junge, and Hughes(2024)}]{pan2024vision}
Pan, C.; Junge, K.; and Hughes, J. 2024.
\newblock Vision-language-action model and diffusion policy switching enables dexterous control of an anthropomorphic hand.
\newblock \emph{arXiv preprint arXiv:2410.14022}.

\bibitem[{Peng et~al.(2019)Peng, Chang, Zhang, Abbeel, and Levine}]{peng2019mcp}
Peng, X.~B.; Chang, M.; Zhang, G.; Abbeel, P.; and Levine, S. 2019.
\newblock Mcp: Learning composable hierarchical control with multiplicative compositional policies.
\newblock \emph{Advances in neural information processing systems}, 32.

\bibitem[{Qi et~al.(2017)Qi, Yi, Su, and Guibas}]{qi2017pointnet++}
Qi, C.~R.; Yi, L.; Su, H.; and Guibas, L.~J. 2017.
\newblock Pointnet++: Deep hierarchical feature learning on point sets in a metric space.
\newblock \emph{Advances in neural information processing systems}, 30.

\bibitem[{Qian et~al.(2014)Qian, Sun, Wei, Tang, and Sun}]{qian2014realtime}
Qian, C.; Sun, X.; Wei, Y.; Tang, X.; and Sun, J. 2014.
\newblock Realtime and robust hand tracking from depth.
\newblock In \emph{Proceedings of the IEEE conference on computer vision and pattern recognition}, 1106--1113.

\bibitem[{Rajeswaran et~al.(2017)Rajeswaran, Kumar, Gupta, Vezzani, Schulman, Todorov, and Levine}]{rajeswaran2017learning}
Rajeswaran, A.; Kumar, V.; Gupta, A.; Vezzani, G.; Schulman, J.; Todorov, E.; and Levine, S. 2017.
\newblock Learning complex dexterous manipulation with deep reinforcement learning and demonstrations.
\newblock \emph{arXiv preprint arXiv:1709.10087}.

\bibitem[{Shao et~al.(2020)Shao, Ferreira, Jorda, Nambiar, Luo, Solowjow, Ojea, Khatib, and Bohg}]{shao2020unigrasp}
Shao, L.; Ferreira, F.; Jorda, M.; Nambiar, V.; Luo, J.; Solowjow, E.; Ojea, J.~A.; Khatib, O.; and Bohg, J. 2020.
\newblock Unigrasp: Learning a unified model to grasp with multifingered robotic hands.
\newblock \emph{IEEE Robotics and Automation Letters}, 5(2): 2286--2293.

\bibitem[{She et~al.(2022)She, Hu, Xu, Liu, Xu, and Huang}]{she2022learning}
She, Q.; Hu, R.; Xu, J.; Liu, M.; Xu, K.; and Huang, H. 2022.
\newblock Learning high-DOF reaching-and-grasping via dynamic representation of gripper-object interaction.
\newblock \emph{arXiv preprint arXiv:2204.13998}.

\bibitem[{Wan et~al.(2023)Wan, Geng, Liu, Shan, Yang, Yi, and Wang}]{wan2023unidexgrasp++}
Wan, W.; Geng, H.; Liu, Y.; Shan, Z.; Yang, Y.; Yi, L.; and Wang, H. 2023.
\newblock Unidexgrasp++: Improving dexterous grasping policy learning via geometry-aware curriculum and iterative generalist-specialist learning.
\newblock In \emph{Proceedings of the IEEE/CVF International Conference on Computer Vision}, 3891--3902.

\bibitem[{Wang et~al.(2023)Wang, Zhang, Chen, Xu, Li, Liu, and Wang}]{wang2023dexgraspnet}
Wang, R.; Zhang, J.; Chen, J.; Xu, Y.; Li, P.; Liu, T.; and Wang, H. 2023.
\newblock Dexgraspnet: A large-scale robotic dexterous grasp dataset for general objects based on simulation.
\newblock In \emph{2023 IEEE International Conference on Robotics and Automation (ICRA)}, 11359--11366. IEEE.

\bibitem[{Wu, Guo, and Liu(2022)}]{wu2022learning}
Wu, A.; Guo, M.; and Liu, C.~K. 2022.
\newblock Learning diverse and physically feasible dexterous grasps with generative model and bilevel optimization.
\newblock \emph{arXiv preprint arXiv:2207.00195}.

\bibitem[{Wu, Wang, and Wang(2023)}]{wu2023learning}
Wu, Y.-H.; Wang, J.; and Wang, X. 2023.
\newblock Learning generalizable dexterous manipulation from human grasp affordance.
\newblock In \emph{Conference on Robot Learning}, 618--629. PMLR.

\bibitem[{Xiang et~al.(2021)Xiang, Zhang, Song, Yu, and Cai}]{xiang2021walk}
Xiang, T.; Zhang, C.; Song, Y.; Yu, J.; and Cai, W. 2021.
\newblock Walk in the cloud: Learning curves for point clouds shape analysis.
\newblock In \emph{Proceedings of the IEEE/CVF international conference on computer vision}, 915--924.

\bibitem[{Xu et~al.(2023)Xu, Wan, Zhang, Liu, Shan, Shen, Wang, Geng, Weng, Chen et~al.}]{xu2023unidexgrasp}
Xu, Y.; Wan, W.; Zhang, J.; Liu, H.; Shan, Z.; Shen, H.; Wang, R.; Geng, H.; Weng, Y.; Chen, J.; et~al. 2023.
\newblock Unidexgrasp: Universal robotic dexterous grasping via learning diverse proposal generation and goal-conditioned policy.
\newblock In \emph{Proceedings of the IEEE/CVF Conference on Computer Vision and Pattern Recognition}, 4737--4746.

\bibitem[{Ye et~al.(2023)Ye, Wang, Huang, Qin, and Wang}]{ye2023learning}
Ye, J.; Wang, J.; Huang, B.; Qin, Y.; and Wang, X. 2023.
\newblock Learning continuous grasping function with a dexterous hand from human demonstrations.
\newblock \emph{IEEE Robotics and Automation Letters}, 8(5): 2882--2889.

\bibitem[{Yu et~al.(2020)Yu, Kumar, Gupta, Levine, Hausman, and Finn}]{yu2020gradient}
Yu, T.; Kumar, S.; Gupta, A.; Levine, S.; Hausman, K.; and Finn, C. 2020.
\newblock Gradient surgery for multi-task learning.
\newblock \emph{Advances in neural information processing systems}, 33: 5824--5836.

\bibitem[{Yu et~al.(2022)Yu, Tang, Rao, Huang, Zhou, and Lu}]{yu2022point}
Yu, X.; Tang, L.; Rao, Y.; Huang, T.; Zhou, J.; and Lu, J. 2022.
\newblock Point-bert: Pre-training 3d point cloud transformers with masked point modeling.
\newblock In \emph{Proceedings of the IEEE/CVF conference on computer vision and pattern recognition}, 19313--19322.

\bibitem[{Zimmermann et~al.(2019)Zimmermann, Ceylan, Yang, Russell, Argus, and Brox}]{zimmermann2019freihand}
Zimmermann, C.; Ceylan, D.; Yang, J.; Russell, B.; Argus, M.; and Brox, T. 2019.
\newblock Freihand: A dataset for markerless capture of hand pose and shape from single rgb images.
\newblock In \emph{Proceedings of the IEEE/CVF international conference on computer vision}, 813--822.

\end{thebibliography}

\end{document}